\documentclass{article}

\usepackage{PRIMEarxiv}

\usepackage[utf8]{inputenc}
\usepackage[T1]{fontenc}
\usepackage{natbib}
\usepackage{hyperref}
\usepackage{url}
\usepackage{booktabs}
\usepackage{amsmath}
\usepackage{amsfonts}
\usepackage{amssymb}
\usepackage{nicefrac}
\usepackage{microtype}
\usepackage{graphicx}
\usepackage{placeins}
\usepackage{float}

\title{\includegraphics[width=1.0\textwidth]{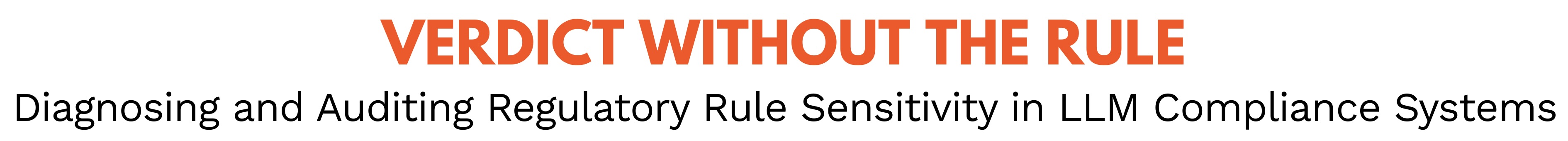}}

\author{
  Saisab Sadhu, Aadit Sengupta\textsuperscript{*} \\
  Vinay Kumar Sankarapu, Pratinav Seth \\
  \affiliation{Lexsi Labs} \\
  \small{\textbf{Correspondence:} \href{mailto:saisab.sadhu@lexsi.ai}{saisab.sadhu@lexsi.ai}}
}

\setabstract{%
Large language model compliance systems are deployed on the assumption
that a verdict depends on the regulatory rule it is given. We test this
directly across five models and 20 regulatory and platform-policy
domains: delete, swap, or negate the governing rule while holding the
case fixed, and check whether the verdict changes (OCS) or the model's
internal representation of compliance shifts at all (ICS-delta). Neither
moves much: models' verdicts are often invariant to substantial
perturbations of the supplied rule, and the guard model, evaluated here
under a custom-rule adaptation of its native taxonomy, is the least
rule-sensitive and least accurate of the five, barely above chance
(51\%, versus 90--92\% for general-purpose models). This reflects easy
cases more than blanket neglect: on cases where deleting the rule
changes a previously correct model prediction, models do track it
closely. Neither better prompting nor direct intervention on the
model's internal representations closes this gap. Accuracy alone does
not establish that a compliance verdict is grounded in the supplied
rule.
}

\setkeywords{LLM compliance systems $\cdot$ rule sensitivity $\cdot$ counterfactual auditing $\cdot$ activation probing $\cdot$ regulatory NLP}

\runningtitle{Verdict Without the Rule}

\begin{document}
\maketitle
{\renewcommand{\thefootnote}{\fnsymbol{footnote}}\footnotetext[1]{Work done while at Lexsi Labs.}}

\section{Introduction}
\label{sec:intro}

Compliance systems built on large language models are already in
production, making legally consequential determinations under the
assumption that they process the regulatory rule they are given.
Standard evaluation measures accuracy against a labeled benchmark
without checking whether the rule was used to produce the verdict at
all. A verdict that survives deleting the rule entirely is not more
defensible for having scored well on such a benchmark: if the rule made
no difference, the system's accuracy reflects surface features of the
case, or patterns learned in training, and calling that a compliance
determination is not warranted. Accuracy alone cannot make this
distinction: a correct verdict reached without consulting the supplied
rule looks identical to one that required it.

This paper tests whether that assumption holds, from two independent
directions. Behaviorally, we ask whether a compliance system's
verdict changes when the governing rule is deleted, swapped for a
different domain's, or has its mandatory obligations negated, holding
the document fixed. Representationally, we ask whether the same
manipulations move the compliance-predictive direction in the model's
residual stream at all, independent of what it ultimately outputs. These
two tests answer different questions. A verdict could stay the same
because the model suppresses the rule's influence right before
answering, even while still representing it internally, or because the
representation we probe never reflects the rule at all. The behavioral
test alone cannot tell these two cases apart; the representational test
can.
(1) \textbf{RQ1}: do LLM compliance verdicts change when the governing
rule is deleted, domain-swapped, or obligation-negated? (2)
\textbf{RQ2}: does the compliance-predictive residual-stream direction
shift under the same rule perturbations? (3) \textbf{RQ3}: do targeted
interventions, rule-grounded prompting and activation steering, recover
accuracy on the cases showing the least rule sensitivity? (4) \textbf{RQ4}: which textual
properties of a rule make a regulatory domain more or less rule-sensitive?

\begin{figure}[t]
\centering
\includegraphics[width=0.99\textwidth]{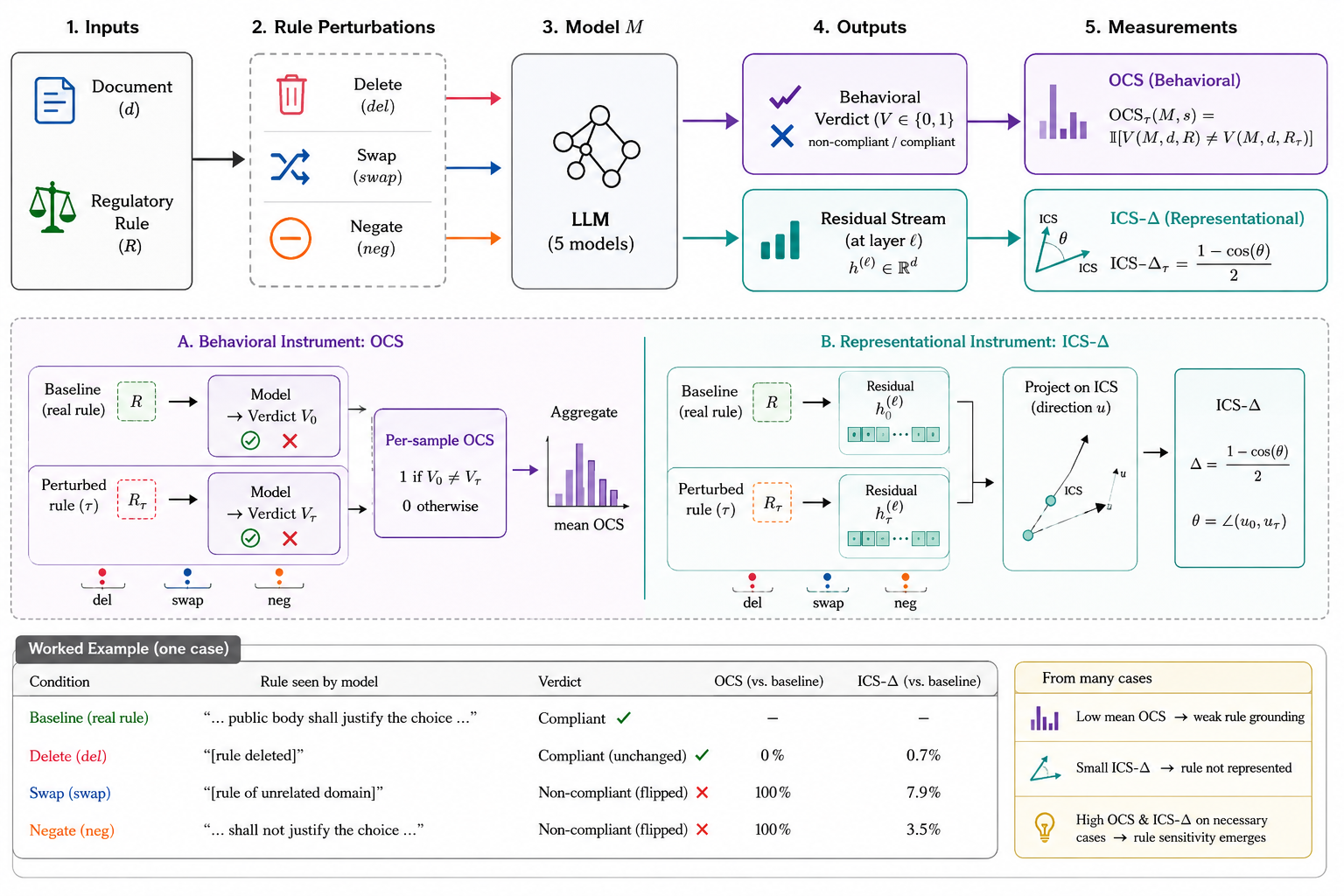}
\caption{The OCS/ICS-delta pipeline (top), each instrument's logic
(middle), and both applied to one real case (bottom): swapping the rule
or negating its obligation flips this model's verdict, but its
compliance-predictive activation moves by no more than 7.9\% of baseline
magnitude. The cosine/angle notation is a conceptual simplification;
Sections~\ref{sec:ocs}--\ref{sec:ics-delta} give the exact definitions
used throughout.}
\label{fig:concept}
\end{figure}

This paper makes four contributions: (1) \textbf{OCS}, a reference-free,
black-box behavioral metric for counterfactual rule dependence, with a
per-sample formulation that identifies specific low-rule-sensitivity
cases rather than only a domain aggregate; (2) \textbf{ICS-delta}, a
representational, perturbation-based probe for whether a
compliance-predictive direction encodes the rule it conditions on,
rather than only whether it predicts the correct label; (3) a
targeted-intervention test that uses OCS to identify the cases where a
system shows the least rule sensitivity, then checks whether
prompting- or representation-level intervention recovers accuracy on
exactly those cases, reporting the outcome as a finding in itself; and
(4) a linguistic account of what makes regulatory text more or less
legible to current systems, from five textual features.

\section{Related Work}
\label{sec:related}

LLM-based compliance systems now span many regulatory settings:
FinGuard \citep{finguard2026}, ComplianceNLP \citep{compliancenlp2026},
GraphCompliance \citep{graphcompliance2025}, LegiLM
\citep{legilm2024}, \citet{judgingbytherules2026}, RegOps-Bench
\citep{regops2026}, and CLAUSE \citep{clause2026}. Each is evaluated on
classification, generation, or retrieval accuracy, not on whether a
verdict is grounded in the rule supplied. \citet{notreadyforthebench2025}
comes closest, showing LLM legal interpretations are unstable across prompt
variations without diagnosing why; OCS and ICS-delta together offer a
behavioral and representational test for that instability. RuleSafe-VL
\citep{rulesafevl2026} makes a related point in vision-language
moderation: final-label accuracy reveals little about whether a model
applied the underlying rule structure, our motivating concern.

Purpose-built safety and policy classifiers
\citep{inan2023llamaguard,han2024wildguard,zhao2025qwen3guard,%
mazeika2024harmbench} run as an additional forward pass, evaluated
almost exclusively on accuracy; DynaGuard \citep{dynaguard2026} and
CourtGuard \citep{courtguard2026} move toward dynamic, inference-time
policy conditioning, a different architectural answer to the problem
OCS diagnoses behaviorally. A parallel faithfulness literature asks
whether stated justifications reflect the computation behind them
\citep{jacovi2020faithfully,lanham2023measuring}; OCS is narrower and
purely behavioral, so it needs no explanation and applies to bare-label
guard models, while ICS-delta brings the same counterfactual logic to
internal representations. Diagnosing reliance on spurious heuristics is
established in NLI \citep{mccoy2019right}; we apply this logic to
compliance. Legal-understanding benchmarks \citep{chalkidis2022lexglue}
formalize legal reasoning without testing rule grounding, and a
separate line monitors violations or conflicts directly from internal
activations or logits
\citep{rachmil2025whitening,gavel2026,refusalresidue2026,highstakes2025,sadhu2026rag},
sharing ICS-delta's representational framing but asking only whether a
violation or conflict signal is present at all, not whether it encodes
the specific rule supplied. Most directly related is LPG \citep{lpg2026}, which removes
the violated policy clause and checks whether the verdict shifts to
safe, and \citet{sadhu2026compliancedetectorsreadaudit}, whose ICS
probe we build on (Section~\ref{sec:ics-delta}) and who test a related
but coarser question: whether deleting, shuffling, or substituting a
rule changes aggregate classifier AUROC across guard models and
activation probes. ICS-delta differs in granularity and construct: it
measures the per-sample magnitude of representational shift under the
same perturbations, rather than an aggregate accuracy change, and we
evaluate it jointly against OCS's independent behavioral counterfactual
rather than in isolation. We instead run a systematic multi-domain
audit, pair behavioral with activation-level sensitivity analysis, and
use a rule-necessity control to expose two accuracy regimes
single-verdict counterfactuals cannot separate.

\section{Methodology}
\label{sec:methodology}
We test rule dependence with two complementary instruments: OCS asks
whether a system's verdict changes when the governing rule is
perturbed, and ICS-delta asks whether the model's internal,
compliance-predictive representation shifts under the same
perturbations. Both apply identical rule-level manipulations
(Section~\ref{sec:conditions}) for directly comparable results.

\subsection{OCS: Behavioral Rule Sensitivity}
\label{sec:ocs}
\textbf{OCS (Output Compliance Sensitivity)} measures how often a
compliance system's verdict changes when the governing rule is deleted,
swapped, or negated, holding the case fixed. A high OCS means the system
reacts to what the rule says; a low OCS means it largely does not.

Formally: let a compliance
system $M$ take as input a document $d$ and a regulatory
rule $R$ and produce a verdict $V \in \{0, 1\}$ (non-compliant / compliant).
Let $T = \{\text{del}, \text{swap}, \text{neg}\}$ be the perturbation
conditions defined below, and write $R_\tau = \mathrm{alter}(R, \tau)$
for the rule after applying condition $\tau$. For a condition $\tau$ and
sample $s = (d, R)$:
\[
\scalebox{0.95}{$\displaystyle
\mathrm{OCS}_\tau(M, s) =
\begin{cases}
1 & V(M, d, R) \neq V(M, d, R_\tau) \\
0 & \text{otherwise}
\end{cases}$}
\]
Since $\mathrm{OCS}_\tau$ is a 0/1 indicator, every quantity built from
it by averaging is bounded in $[0,1]$: 0 means the verdict never changes
under that perturbation, 1 means it always changes, and domain-level and
aggregate OCS follow by averaging over samples and conditions. A
\textbf{per-sample} aggregate, $\mathrm{OCS}_{\text{sample}}(M, s)$,
averages over the applicable conditions (negation excluded for samples
with no mandatory modal, Section~\ref{sec:conditions}). It defines two
partitions used in Section~\ref{sec:interventions}: LOW-OCS (unchanged
under every applicable condition) and HIGH-OCS (changed under at least
two of three). Both partitions are defined on the \textbf{primary
model}, Llama-3.3-70B-Instruct, and applied identically to every other
model.
Under random, rule-unrelated verdict-flipping and an unbiased marginal,
expected OCS is $\approx 0.5$; Section~\ref{sec:results-audit} replaces
this with an empirical null accounting for each model's response bias.

\paragraph{Interpretive scope of OCS.} OCS is a diagnostic
proxy for counterfactual rule dependence, not a correctness measure: low
OCS means verdicts are largely invariant to rule manipulations, but a
verdict can be correct without the rule having been consulted, so
invariance alone does not establish incorrect reasoning.
Section~\ref{sec:necessity-control}'s rule-necessity control addresses
this directly. OCS also measures counterfactual sensitivity, not
whether a changed verdict would be legally required under the perturbed
rule; low OCS by itself should not be read as proof that a model
ignored the rule, only that its verdict did not change under our
perturbations.

\subsubsection{Perturbation Conditions}
\label{sec:conditions}
Three conditions, applied to the rule field only with the document held
fixed, none requiring an external API call:
(1) \textbf{OCS-del}: the rule is deleted (empty string), the most
basic test, since a verdict that survives no rule at all is not, in
any operational sense, applying it. (2) \textbf{OCS-swap}: the rule is
replaced with the canonical (most frequent) rule text of a
uniformly-sampled \emph{different} regulatory domain, testing semantic
and domain specificity. (3) \textbf{OCS-neg}: the rule's mandatory
modals (``must''/``shall'') are negated (``must not''/``shall
not''); permissive and recommendation modals (``may''/``can''/``should'')
are left untouched. Applied per sample, not per domain: samples with no
mandatory modal are excluded from OCS-neg rather than counted as
unchanged. Per-domain modal coverage ranges 0.0--60.7\%
(Appendix~\ref{app:domains}, Table~\ref{tab:domains}).

\subsection{ICS-delta: Representational Grounding}
\label{sec:ics-delta}
\textbf{ICS (Internal Compliance Score)} \citep{sadhu2026compliancedetectorsreadaudit} is a
training-free probe: a normalized mean-difference direction between
compliant and non-compliant activation clusters at a per-model,
per-domain \textbf{anchor layer}, selected by argmax AUROC on a
held-out validation split disjoint from the direction-fitting split.
ICS-delta then asks whether the activation's projection onto this fixed
direction changes under the same rule perturbations used for OCS.

\subsubsection{ICS and ICS-delta}
For
a new (document, rule) pair, ICS is the projection of the last-token
residual-stream activation onto this direction: a real-valued scalar,
larger for stronger alignment with the compliant cluster. The probe's
fitting procedure and validation are described in
\citet{sadhu2026compliancedetectorsreadaudit}; ICS-delta is our own
construction on top of this probe, applying the same rule-perturbation
logic used for OCS to the direction it defines. For a sample $s$ and
condition $\tau$, with $R_\tau
= \mathrm{alter}(R, \tau)$ as in Section~\ref{sec:ocs}:
\[
\scalebox{0.9}{$\displaystyle
\text{ICS-delta}_\tau(M, s) = \left| \text{ICS}(M, d, R) - \text{ICS}(M, d, R_\tau) \right|$}
\]
computed with the document and all other case text held fixed: only the
rule field changes, mirroring the OCS perturbation exactly but scored
via the activation probe rather than an elicited verdict. Unlike OCS,
ICS-delta is non-negative by construction but has no fixed upper bound,
since ICS itself is unbounded, so we report it as a fraction of each
model's own mean baseline $|$ICS$|$ magnitude rather than as a raw
number not comparable across models or domains.

\subsubsection{Interpreting Small ICS-delta}
A low OCS score alone is compatible with two explanations, with
different practical implications: the rule may be represented internally
but its influence suppressed by the time a verdict is produced (an
inference-time fix might recover grounding), or the rule may be absent
from the internal representation entirely (none will).

Rule perturbations produce changes equal to 12.8--22.9\% of mean
baseline $|$ICS$|$ in this particular probe direction, pooled across
conditions, for all four Tier-1 models: 12.8\% (Llama-3.1-8B), 21.0\%
(Llama-3.3-70B), 22.9\% (Mistral-7B-v0.3), 16.7\% (Qwen2.5-7B); OCS-neg
produces the smallest change in projection for every model%
, whereas pooled over models OCS-del changes verdicts least (Section~\ref{sec:results-audit};
full numbers and figure in Appendix~\ref{app:convergent-validity}).
This suggests limited sensitivity along the measured
compliance-predictive direction: if the projection onto this direction
changed substantially with the rule, ICS-delta should be a much larger
fraction of baseline than the 13--23\% observed. It does not establish
that rule information is absent elsewhere in the representation. We do
not claim rule content is absent from the residual stream generally,
only that it weakly affects this particular compliance-predictive
direction; it could still be encoded in orthogonal directions, at other
layers, or at rule-token positions rather than the final token. Nor do
we claim its influence on this direction is fully absent, since the
delta is not exactly zero. Neither intervention tested here (Section~\ref{sec:interventions})
reliably improves accuracy on the LOW-OCS cases; this rules out these
specific prompt- and steering-based fixes, but not inference-time
recovery methods more generally.

\label{sec:convergent-validity}
As a methodological check, we also Spearman-correlate per-domain
OCS-agg and ICS-delta-agg rankings for each Tier-1 model: none show
significant convergence (Appendix~\ref{app:convergent-validity}),
suggesting OCS and ICS-delta probe two related, partially independent
properties (Section~\ref{sec:discussion} returns to this point).

\section{Experimental Setup}
\label{sec:setup}

\subsection{Dataset and Domains}
All experiments use OmniCompliance-100K \citep{hu2026omnicompliance},
with labels derived from regulatory enforcement text (COMPLIES/VIOLATES)
rather than model self-labeling. We use a fixed 20-domain subset
spanning data protection, AI safety, finance, cybersecurity, education,
platform policy, and fundamental rights (full list and per-domain sizes
in Appendix~\ref{app:domains}, Table~\ref{tab:domains}).

\subsection{Sampling}
We sample 200 cases per domain per model (100 COMPLIES/100 VIOLATES
where the domain supports it, the natural ratio otherwise for the four
smallest domains), stratified and seeded identically (seed 42) across
all five models so every model sees the same underlying cases: 12{,}897
(case, condition) generations per model, 64{,}485 total, parseable-verdict
rate above 99.5\% for every model (Appendix~\ref{app:domains}).

\subsection{Models}
\label{sec:models}
Five models, run locally with no closed-source or API access:
Llama-3.1-8B-Instruct and Llama-3.3-70B-Instruct (bf16/int8
respectively), Qwen2.5-7B-Instruct, Mistral-7B-Instruct-v0.3, and
Llama-Guard-3-8B \citep{inan2023llamaguard}, prompted with a custom
compliance-policy category substituted for its fixed content-safety
taxonomy (Appendix~\ref{app:llamaguard}) rather than through its
packaged chat template, which has no field for an arbitrary rule.

Llama-3.3-70B-Instruct, the largest and most capable general-purpose
model in the roster, is the \textbf{primary model} used wherever a
single representative model is needed, e.g.\ the generalization check in
Section~\ref{sec:generalization}. The four general-purpose models are
\textbf{Tier-1}: the models for which we fit an ICS probe direction
(Section~\ref{sec:ics-delta}) and report ICS-delta and
convergent-validity results. Llama-Guard-3-8B is evaluated behaviorally
(OCS) throughout but excluded from these representational analyses,
since its architecture and task format differ too much from the other
four for a shared probe direction to mean the same thing. A sixth
model, WildGuard \citep{han2024wildguard}, was excluded outright: its
fixed subcategory taxonomy takes no rule-conditioning input, so it
cannot produce a meaningful verdict without a contrived reformulation.

\section{Results: Rule Sensitivity Audit}
\label{sec:results-audit}

\subsection{OCS Is Uniformly Low}
\begin{figure}[t]
\centering
\includegraphics[width=0.4617\columnwidth]{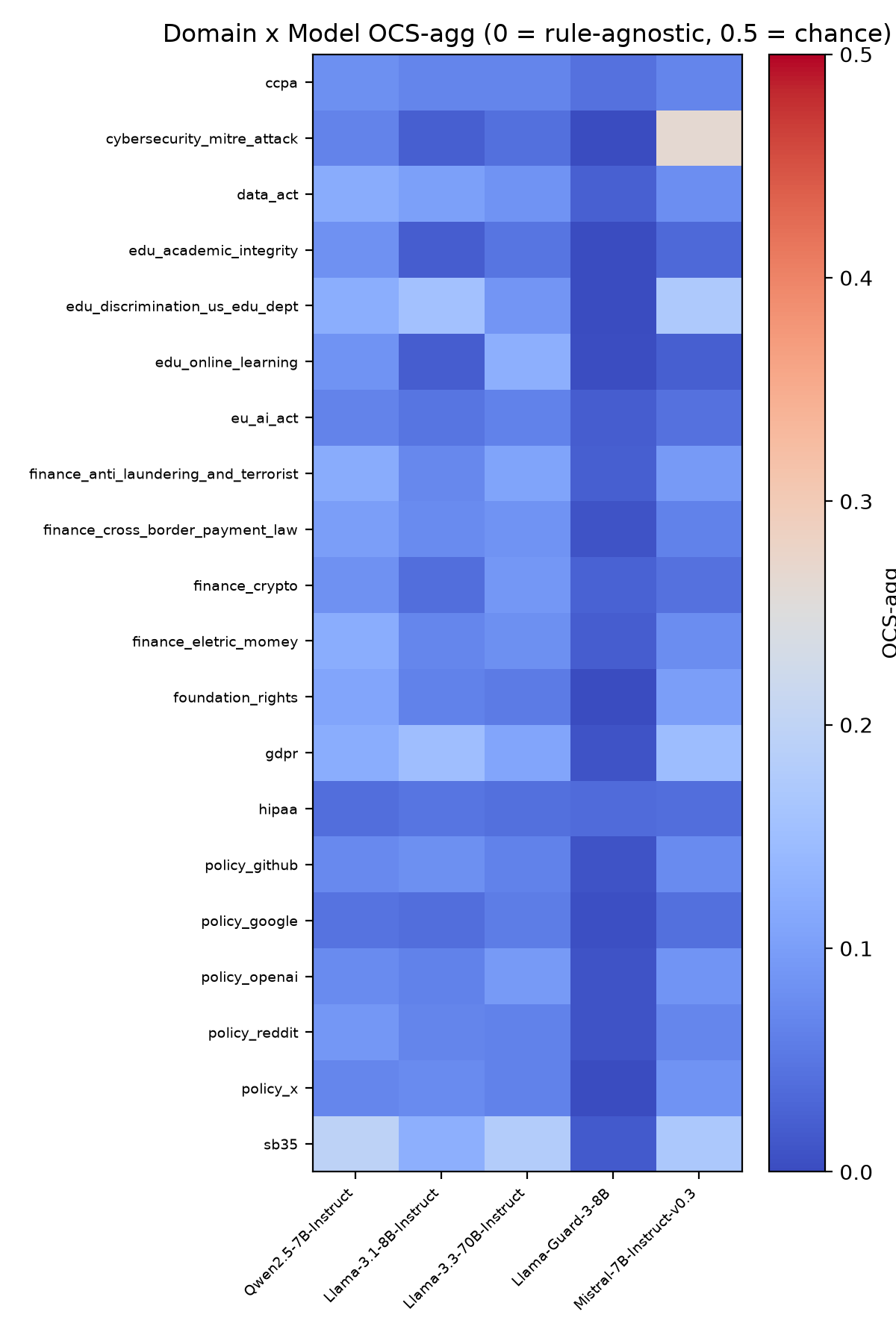}
\caption{Domain $\times$ model OCS-agg. Nearly every cell is cold (low
rule sensitivity); the one visibly warm cell (Mistral-7B-v0.3 $\times$
cybersecurity\_mitre\_attack, 0.265) is the single highest value in the
dataset.}
\label{fig:heatmap}
\end{figure}

Table~\ref{tab:model-summary} reports OCS-agg and baseline accuracy per
model, pooled across all 20 domains. OCS-agg is well below the
$\approx 0.5$ independence reference for every model, ranging from
0.013 (Llama-Guard-3-8B) to 0.094 (Qwen2.5-7B-Instruct), with mean 0.069
across all 100 (model, domain) cells. Figure~\ref{fig:heatmap} shows the
full domain $\times$ model grid: nearly every cell is cold. Broken down
by perturbation type and pooled across all models and domains, OCS-del
(rule deleted) is the lowest of the three at 0.053, below OCS-neg
(0.070) and OCS-swap (0.082). Verdicts are least likely to change when
the rule is removed entirely, and somewhat more likely, though still
rare, when it is replaced with an unrelated domain's rule or has its
obligations reversed.

\begin{table}[t]
\centering
\caption{OCS by perturbation condition, pooled across all 20 domains
(200 cases/domain, seed 42). Per-domain values in Figure~\ref{fig:heatmap}.}
\vspace{3mm}
\label{tab:model-summary}
\footnotesize
\setlength{\tabcolsep}{8pt}
\renewcommand{\arraystretch}{1.2}
\begin{tabular}{@{}lcccc@{}}
\toprule
Model & OCS-del & OCS-swap & OCS-neg & OCS-agg \\
\midrule
Qwen2.5-7B & 0.052 & 0.118 & 0.135 & 0.094 \\
Llama-3.1-8B & 0.064 & 0.079 & 0.060 & 0.070 \\
Llama-3.3-70B & 0.061 & 0.095 & 0.084 & 0.081 \\
Llama-Guard-3-8B & 0.014 & 0.015 & 0.009 & 0.013 \\
Mistral-7B-v0.3 & 0.076 & 0.103 & 0.065 & 0.089 \\
\bottomrule
\end{tabular}
\end{table}

Table~\ref{tab:accuracy-perturbation} (Appendix~\ref{app:domains})
decomposes baseline accuracy by condition, derived per sample from each
condition's verdict-flip indicator against the gold label. Deletion
accuracy, the closest proxy for case-content-only performance, stays
within 4 points of baseline for every general-purpose model, slightly
\emph{above} it for Qwen2.5-7B and Mistral-7B-v0.3. Llama-Guard-3-8B
stays flat at chance throughout. Swap and negation drop accuracy
further, to 79.5--89.7\%, tracking each model's higher OCS-swap/OCS-neg
rate (Table~\ref{tab:model-summary}).

Domain-level variance exists, but it is dominated by model-specific
effects. The one visibly warm cell in Figure~\ref{fig:heatmap}
(Mistral-7B-v0.3 $\times$ \texttt{cybersecurity\_mitre\_attack}, OCS-agg
0.265) is not replicated by any other model on that domain. Across
models, \texttt{sb35} (a US state AI-safety statute) shows the highest
domain-mean OCS (0.138), the closest this dataset comes to a
model-general effect (per-condition breakdown in
Figure~\ref{fig:perturbation-bars}, Appendix~\ref{app:domains}).
Section~\ref{sec:linguistic}, next, asks what drives this variance.
OCS-agg also shows a weak, marginal positive correlation with baseline
accuracy across the 100 (model, domain) cells ($\rho = 0.190$,
$p = 0.058$; Appendix~\ref{app:accuracy-predictor}).

\subsection{Linguistic Regression}
\label{sec:linguistic}
Five features are extracted from every rule text in each domain, not
only a single canonical rule: exception-clause density (ECD),
cross-reference count (CRC), mandatory-modal ratio (MMR), mean
dependency-tree depth (MDD), and Flesch-Kincaid grade level (FKGL). Each
is correlated (Spearman) against each domain's mean OCS-agg across all
five models, and jointly in a standardized OLS regression
(Table~\ref{tab:regression}; joint coefficients in
Figure~\ref{fig:regression-coefs}, Appendix~\ref{app:linguistic}),
with every $p$-value corrected across the five tests via
Benjamini-Hochberg (BH), which controls the expected proportion of false
positives among features called significant.

Exception-clause density and cross-reference count, two features with
strong theoretical motivation, do not reach significance after
correction (ECD $\rho=0.041$, $p_{BH}=0.86$; CRC $\rho=0.241$,
$p_{BH}=0.38$). The one feature that survives BH correction is \textbf{mean dependency-tree depth} (MDD): $\rho=0.570$,
$p_{BH}=0.044$%
. MMR
($\rho=0.405$) and FKGL ($\rho=0.448$) are moderate but not significant
after correction.

The joint OLS model explains under half the variance ($R^2=0.484$,
adjusted $R^2=0.285$) and does not itself reach significance at the
conventional threshold ($F=2.44$, $p=0.091$, $n=19$ domains;
\texttt{edu\_academic\_integrity} is dropped for undefined MMR, having
zero modal verbs of any kind). MDD's correlation is positive: higher
dependency-tree depth predicts \emph{higher} OCS, i.e., more rule
sensitivity, not less. This supports the view that a rule's syntactic
complexity predicts rule sensitivity better than the surface features
(exception clauses, cross-references) considered here, though the
overall regression is suggestive rather than decisive. (Full per-feature
table in Appendix~\ref{app:linguistic}, Table~\ref{tab:regression}.)

\subsection{OCS vs.\ an Empirical Null}
\label{sec:empirical-null}
The $\approx 0.5$ reference above assumes an unbiased verdict marginal,
which does not hold for every model, least of all Llama-Guard-3-8B. We
construct a stricter, model-specific null: holding each model's own
empirical baseline-verdict rate $p$ and altered-verdict rate $q$ fixed
(pooled across domains and conditions), the flip rate expected under
independence is $p(1-q) + q(1-p)$, validated by a permutation test that
randomly re-pairs altered verdicts to baseline verdicts within each
domain 10,000 times, breaking any true dependence while preserving
domain composition.

For the four general-purpose models, this bias-corrected null
(0.48--0.50) sits close to the naive 0.5 reference, and observed OCS-agg
is far below it in every one of 10,000 permutations ($p<0.0001$ for all
four; full per-model null, observed value, and $p$ in
Appendix~\ref{app:necessity}, Table~\ref{tab:permutation-null}).
Llama-Guard-3-8B differs: its response bias is extreme
($p=0.985$, $q=0.996$; it answers ``compliant'' almost regardless of
condition), dropping its bias-corrected null to $0.019$, yet its
observed OCS ($0.013$) is still significantly lower ($p<0.0001$),
reflecting more rule-insensitivity than its own response bias alone
would predict.

\subsection{Rule-Necessity Control}
\label{sec:necessity-control}
A skeptical reading of the low aggregate OCS above is that case content
alone may already determine the correct verdict for most samples,
making the rule decision-irrelevant regardless of whether a model
attends to it. We test this directly: for each model, we isolate cases
where the baseline verdict (real rule present) is correct but the
verdict under rule deletion is incorrect, i.e., where removing the rule
demonstrably changes a correct model prediction into an incorrect one.
Here ``rule-necessary'' is operational: it denotes this correct-then-incorrect
pattern, not an independently annotated claim that the legal task
itself requires the rule. This subset is model-specific, ranging from
94 (Qwen2.5-7B-Instruct) to 187 (Llama-3.3-70B-Instruct) cases among the general-purpose models (49 for Llama-Guard-3-8B); only 6
sampled cases are rule-necessary for all four general-purpose models at
once.

The result is the opposite of a confound. Restricted to this subset,
OCS-swap rises sharply for every model (0.73--1.00, vs.\ 0.08--0.12
pooled) and, for one of four general-purpose models, significantly
\emph{exceeds} its own bias-corrected null after Bonferroni correction at
the family threshold of $0.005$ across ten tests
(Llama-3.1-8B-Instruct $p<0.0001$); Mistral-7B-v0.3 is borderline
($p=0.006$, threshold $0.005$; full table in Appendix~\ref{app:necessity},
Table~\ref{tab:necessity-appendix}). On these deletion-critical cases,
models become substantially more sensitive to rule swaps; on the
majority that are not deletion-critical, accuracy on a labeled
benchmark gives no way to determine which regime produced the verdict.

\subsection{Guard-Model Results}
\label{sec:guard-finding}
Llama-Guard-3-8B, a safety guard model adapted here to arbitrary
regulatory-rule conditioning outside its native content-safety
taxonomy, has the lowest OCS-agg of all five models (0.013, roughly a
sixth of the general-purpose models' 0.070--0.094) and a baseline
accuracy of only 51.1\%, barely above chance, against 90.3--91.9\% for
every general-purpose model on the identical cases. Under our
custom-rule adaptation, which substitutes regulatory text into a slot
its fixed 14-category taxonomy was not trained for
(Appendix~\ref{app:llamaguard}), it shows both the lowest rule
sensitivity and lowest accuracy, consistent with task-format mismatch
rather than a general claim about compliance-specific training. The
four general-purpose models are the paper's primary evidence; the guard
model is a complementary case, a different failure mode rather than a
more severe version of the same one: it has no learned representation
of what GDPR compliance looks like in a slot built for one of its own
14 labels. The general-purpose models show low rule sensitivity at high
accuracy: they reach a correct verdict through some route other than
the rule. Llama-Guard-3-8B shows low rule sensitivity at chance
accuracy: it does not reach a correct verdict at all. One substitutes a
different signal for the rule; the other is performing a task it was
never fit to do.

\section{Cross-Dataset Generalization}
\label{sec:generalization}
To check whether the OCS protocol and the low-OCS finding are specific
to OmniCompliance-100K or hold on independently constructed benchmarks,
we run the identical three-condition OCS sweep (Llama-3.3-70B-Instruct
only; no ICS-delta, interventions, or regression on these) on two
further datasets. \textbf{LegalBench} \citep{legalbench2023guha}: 887
samples, 272 from \texttt{sara\_entailment} (a statutory tax-entailment
task where the rule is a specific numbered provision and the document is
facts plus the numeric claim tested) and 615 from
\texttt{privacy\_policy\_entailment}, the closest available
OPP-115-adjacent rule-application task with a binary label (the literal
\texttt{opp115\_*} configs are multi-label span-category classification
with no rule field and do not fit the OCS protocol).
\textbf{ContractNLI} \citep{koreeda2021contractnli}: 400 samples
stratified 200 entailment and 200 contradiction (neutral dropped),
hypothesis as rule, contract excerpt as document; 387 have a parseable
baseline verdict, the $n$ reported in Table~\ref{tab:generalization}.

Table~\ref{tab:generalization} shows OCS-agg is higher on both external
benchmarks than on OmniCompliance-100K: 0.255 on LegalBench
(%
$3.1\times$) and 0.318 on ContractNLI ($3.9\times$). This argues
against a single, universal degree of low rule sensitivity. LegalBench's
baseline accuracy (49.7\%) is indistinguishable from chance, so part of
its elevated OCS plausibly reflects task difficulty rather than genuine
rule-grounding. ContractNLI's (78.6\%) is well above chance, making its
higher OCS a cleaner case: here the model is both more accurate and
more rule-sensitive. This plausibly owes to format. ContractNLI's
hypothesis-as-rule, contract-excerpt-as-document structure gives every
case a tighter logical link than OmniCompliance-100K's narrative pairs,
so task format, not only model identity, appears to affect how much a
verdict depends on the rule supplied. Unparseable-verdict rates are also
higher on both (1.4\%, 3.3\%) than OmniCompliance-100K's near-zero
rate.

\begin{table}[t]
\centering
\caption{Generalization check, Llama-3.3-70B-Instruct only.
$^\dagger$OmniCompliance-100K row restated from Table~\ref{tab:model-summary} and Table~\ref{tab:accuracy-perturbation}.
OCS stays well under the 0.5 independence reference everywhere but is
not a fixed, dataset-independent property of the model. $n$ counts sampled cases, except for ContractNLI, where it counts the cases with a parseable baseline verdict.}
\vspace{3mm}
\label{tab:generalization}
\footnotesize
\setlength{\tabcolsep}{8pt}
\renewcommand{\arraystretch}{1.2}
\begin{tabular}{@{}lccc@{}}
\toprule
Dataset & OCS-agg & Baseline acc. & $n$ \\
\midrule
OmniCompliance-100K$^\dagger$ & 0.081 & 0.909 & 3{,}946 \\
LegalBench & 0.255 & 0.497 & 887 \\
ContractNLI & 0.318 & 0.786 & 387 \\
\bottomrule
\end{tabular}
\end{table}

Low rule sensitivity is not an artifact of one dataset's rule-text
style: the finding holds (OCS remains well under the independence
reference) on both independently constructed benchmarks, though its
magnitude is not a fixed model property, running substantially higher
outside OmniCompliance-100K. Practitioners should compute OCS on data
resembling their actual deployment distribution rather than relying on
any single portable number.

\section{Interventions}
\label{sec:interventions}

We test two interventions on the LOW-OCS and HIGH-OCS partitions defined
in Section~\ref{sec:ocs}.

\subsection{Rule-Grounded Prompting}
\label{sec:rule-grounded}
A modified prompt requires the model to first quote the specific rule
clause it considers most relevant, then state a verdict, testing whether
making rule engagement explicit recovers grounding that implicit
prompting does not.

Table~\ref{tab:exp3a} (full table in Appendix~\ref{app:steering})
reports $\Delta_{\text{low}}$ and $\Delta_{\text{high}}$, the accuracy
change from standard to rule-grounded prompting on the LOW-OCS and
HIGH-OCS partitions, with McNemar's exact test (the same cases scored
twice, once per prompt, so what matters is whether the two scorings
disagree and in which direction). Only one of ten cells reaches
significance: Qwen2.5-7B-Instruct's LOW-OCS partition, $\Delta=-0.028$
($p<0.0001$): rule-grounded prompting significantly \emph{harms}
accuracy on exactly the cases it was meant to help. Mistral-7B-v0.3 and
Llama-3.3-70B-Instruct's LOW-partition deltas ($-0.010$, $+0.007$) point
opposite ways, but both fall short of significance ($p=0.077$,
$p=0.073$). Llama-3.1-8B and Llama-Guard-3-8B show no effect in either
partition. The one statistically supported finding here is negative: no
model improves significantly, and one significantly backfires.

\subsection{Targeted Activation Steering}
For two of the four models with a fitted ICS direction (Llama-3.3-70B-Instruct,
Llama-3.1-8B-Instruct), we add $\alpha \cdot$ direction to the residual
stream at the anchor layer during generation, where positive $\alpha$
steers the representation toward the compliant cluster and negative
$\alpha$ toward the non-compliant cluster. We sweep
$\alpha \in \{-1, -0.5, 0.5, 1\}$ ($\alpha=0$ is the unsteered baseline),
on the LOW-OCS and HIGH-OCS partitions separately.

\begin{table}[t]
\centering
\caption{Accuracy (\%) by steering $\alpha$, all four (model, partition)
cells. Bold marks the unsteered baseline. $^*p<0.05$, $^{**}p<0.01$
(McNemar's exact test vs.\ the baseline; nominal, uncorrected for the
16 comparisons in this table). Exact $p$-values for every cell in
Appendix~\ref{app:steering}.}
\vspace{3mm}
\label{tab:steering}
\footnotesize
\setlength{\tabcolsep}{6pt}
\renewcommand{\arraystretch}{1.2}
\begin{tabular}{@{}llccccc@{}}
\toprule
Model & Part. & $-1$ & $-0.5$ & $0$ & $+0.5$ & $+1$ \\
\midrule
Llama-3.1-8B & LOW & 92.3$^*$ & 92.5 & \textbf{92.8} & 92.5 & 92.3$^*$ \\
Llama-3.1-8B & HIGH & 79.7 & 82.2 & \textbf{82.2} & 82.7 & 82.2 \\
Llama-3.3-70B & LOW & 91.7$^{**}$ & 91.9 & \textbf{92.1} & 92.3 & 92.3 \\
Llama-3.3-70B & HIGH & 73.3$^*$ & 76.7 & \textbf{79.2} & 79.2 & 78.7 \\
\bottomrule
\end{tabular}
\end{table}

Figure~\ref{fig:steering-sweep} (Appendix~\ref{app:steering}) plots
the same sweep. For Llama-3.1-8B-Instruct, the unsteered baseline is the
single best-performing LOW-OCS setting: every steering direction
reduces accuracy, with the two most extreme cells nominally significant
before correction. For Llama-3.3-70B-Instruct the pattern is
asymmetric: negative steering shows nominally significant harm on both
partitions, while positive steering shows a monotonic-looking,
non-significant improvement on the LOW partition. No setting, at either
scale, significantly beats baseline%
; several negative-steering
comparisons are nominally significant before correcting for the 16
comparisons tested here, so we treat these per-setting results as
exploratory.

The one consistent pattern across both models points the same
direction: steering away from the compliant direction tends to degrade
accuracy on the LOW-OCS partition, while steering toward it never
significantly helps, though these per-setting comparisons are
uncorrected for multiple testing (above). Read with the prompting
result (Section~\ref{sec:rule-grounded}), neither intervention produces
a significant, generalizable recovery on the LOW-OCS cases: whatever
rule-conditioned signal these verdicts are computed from, it is not
enough for a prompt or a steering vector to recover.

\section{Discussion}
\label{sec:discussion}

\paragraph{Audit result and interpretation.} OCS-agg is uniformly low across every model and domain
tested, well below the independence reference. But the rule-necessity
control shows this aggregate conceals two distinct regimes. On cases
where deleting the rule provably disrupts the correct verdict, OCS-swap
rises to 0.73--1.00: models become substantially more sensitive to rule
swaps on these deletion-critical cases. On the majority of cases, where the answer is recoverable
largely from scenario content, verdicts are largely invariant to what
the rule says. Standard accuracy cannot tell which regime produced a
given correct answer.%

\paragraph{The guard model and the two instruments.} The guard model
scores worst on both rule sensitivity and accuracy
(Section~\ref{sec:guard-finding}). If safety-training conferred rule
sensitivity as a side effect, this is where it should appear, and it
does not. The behavioral and representational instruments do not
converge either (Section~\ref{sec:convergent-validity}), suggesting two
separate, related properties, and the interventions offer no clean fix
(Section~\ref{sec:interventions}).

\paragraph{Scope of the evidence.} OCS-del is lower than OCS-swap or
OCS-neg: verdicts are not simply ignoring a fixed prior, since a wrong
rule sometimes clashes with the model's document-based prior, i.e.,
surface sensitivity rather than regulatory content. The intervention
results provide no evidence that these particular inference-time
methods recover the missing sensitivity.

\paragraph{Practical implication.} Accuracy on a labeled benchmark does
not license an inference that verdicts are grounded in the rule
supplied. Our guard-model result suggests this inference needs particular care when an off-the-shelf safety classifier is repurposed to check regulatory rules. OCS is cheap to compute,
needs no gold labels, and could run as a pre-deployment check
independent of a vendor's accuracy claims%
, though, as
Section~\ref{sec:ocs} notes, it flags whether a verdict changed under
perturbation, not whether that change would have been the legally
correct one, so a low score motivates scrutiny rather than settling it.

\section{Conclusion}
We tested, behaviorally and representationally, whether LLM-based
compliance verdicts depend on the rule they are given. Across five
models and 20 domains, many correct verdicts are largely invariant to
rule deletion, substitution, or negation. A rule-necessity control
substantially addresses the simplest objection to this pattern: sensitivity rises
sharply where the rule is outcome-determinative (OCS-swap reaches
0.73--1.00), exposing two regimes standard accuracy cannot distinguish
(full per-model significance testing in Appendix~\ref{app:necessity}).
The guard model, evaluated under our custom-rule adaptation, remains
least rule-sensitive and least accurate of the five, and the two
instruments do not converge.

Two targeted interventions fail to reliably recover accuracy on
LOW-OCS cases, and one measurably worsens it. The strongest predictor
of domain-level rule sensitivity is a rule's syntactic complexity, not
its exception clauses or cross-references. The finding also holds on
LegalBench and ContractNLI, well below the independence reference,
though rule sensitivity there runs 3--4$\times$ higher than on
OmniCompliance-100K, against treating any single OCS number as
portable. None of this proves the pattern permanent or unfixable by
untested methods, and it does not establish that these models never
encode or use regulatory rules. It shows that high benchmark accuracy
can coexist with substantial invariance to changes in the supplied
rule, so accuracy alone is insufficient evidence of rule-grounded
compliance reasoning, particularly for safety classifiers repurposed for compliance.

\section*{Limitations}
\label{sec:limitations}

The standard prompt truncates rule text at 300 characters. This design
means our results characterize rule sensitivity under the evaluated
prompt format, not necessarily sensitivity to complete regulatory
texts: when an operative clause falls beyond the 300-character window,
low OCS can reflect unavailable rule information rather than failure
to use available information. The rule-grounded
prompting intervention simultaneously adds the full rule and changes
reasoning instructions, so the two prompts are not controlled for rule
completeness. The relatively small accuracy drop under deletion
(Table~\ref{tab:accuracy-perturbation}) suggests scenario-side label
signal is present in OmniCompliance-100K, consistent with its
construction methodology. Future work should measure scenario-only
prediction directly and control rule truncation separately.

OCS measures verdict change under perturbation, not whether the change
is legally correct, and it targets only the rule field; case-text
perturbation is a complementary failure mode not addressed here. The
LOW/HIGH-OCS partition is defined on Llama-3.3-70B-Instruct and
transferred to all other models, which assumes LOW-OCS cases are
partly consistent across architectures, an assumption we do not test.
A model whose LOW-OCS cases differ substantially would have its
intervention tested on the wrong subset, undetectably. OCS-neg is
inapplicable to three domains with zero mandatory-modal coverage
(Appendix~\ref{app:domains}), and WildGuard's and Llama-Guard-3-8B's
task-format caveats are in Section~\ref{sec:models}.

ICS-delta is a relative measure along one compliance-predictive
direction at one anchor layer. Rule content may be encoded elsewhere in
the residual stream, and low ICS-delta does not rule that out. The
rule-necessity subsets are small (Appendix~\ref{app:necessity}), leaving
OCS-neg underpowered there, and the generalization check
(Section~\ref{sec:generalization}) uses only one model.
\bibliographystyle{plainnat}
\bibliography{references}

@inproceedings{hu2026omnicompliance,
  title={{OmniCompliance-100K}: A Multi-Domain, Rule-Grounded, Real-World Safety Compliance Dataset},
  author={Hu, Wenbin and Jing, Huihao and Shi, Haochen and Fan, Changxuan and Li, Haoran and Song, Yangqiu},
  booktitle={Findings of the Association for Computational Linguistics: ACL 2026},
  pages={2445--2463},
  year={2026},
  doi={10.18653/v1/2026.findings-acl.115},
  url={https://aclanthology.org/2026.findings-acl.115/}
}

@misc{inan2023llamaguard,
  title={Llama Guard: {LLM}-based Input-Output Safeguard for Human-{AI} Conversations},
  author={Inan, Hakan and Upasani, Kartikeya and Chi, Jianfeng and Rungta, Rashi and Iyer, Krithika and Mao, Yuning and Tontchev, Michael and Hu, Qing and Fuller, Brian and Testuggine, Davide and Khabsa, Madian},
  year={2023},
  eprint={2312.06674},
  archivePrefix={arXiv},
  url={https://arxiv.org/abs/2312.06674}
}

@inproceedings{han2024wildguard,
  title={{WildGuard}: Open One-stop Moderation Tools for Safety Risks, Jailbreaks, and Refusals of {LLMs}},
  author={Han, Seungju and Rao, Kavel and Ettinger, Allyson and Jiang, Liwei and Lin, Bill Yuchen and Lambert, Nathan and Choi, Yejin and Dziri, Nouha},
  booktitle={Advances in Neural Information Processing Systems 37 (Datasets and Benchmarks Track)},
  pages={8093--8131},
  year={2024},
  doi={10.52202/079017-0261}
}

@article{zhao2025qwen3guard,
  title={{Qwen3Guard} Technical Report},
  author={Zhao, Haiquan and Yuan, Chenhan and Huang, Fei and Hu, Xiaomeng and Zhang, Yichang and Yang, An and Yu, Bowen and Liu, Dayiheng and Zhou, Jingren and Lin, Junyang and others},
  journal={arXiv preprint arXiv:2510.14276},
  year={2025}
}

@inproceedings{mazeika2024harmbench,
  title={{HarmBench}: A Standardized Evaluation Framework for Automated Red Teaming and Robust Refusal},
  author={Mazeika, Mantas and Phan, Long and Yin, Xuwang and Zou, Andy and Wang, Zifan and Mu, Norman and Sakhaee, Elham and Li, Nathaniel and Basart, Steven and Li, Bo and Forsyth, David and Hendrycks, Dan},
  booktitle={Proceedings of the 41st International Conference on Machine Learning},
  series={Proceedings of Machine Learning Research},
  volume={235},
  pages={35181--35224},
  year={2024}
}

@inproceedings{jacovi2020faithfully,
title = "Towards Faithfully Interpretable {NLP} Systems: How Should We Define and Evaluate Faithfulness?",
    author = "Jacovi, Alon  and
      Goldberg, Yoav",
    editor = "Jurafsky, Dan  and
      Chai, Joyce  and
      Schluter, Natalie  and
      Tetreault, Joel",
    booktitle = "Proceedings of the 58th Annual Meeting of the Association for Computational Linguistics",
    month = jul,
    year = "2020",
    address = "Online",
    publisher = "Association for Computational Linguistics",
    url = "https://aclanthology.org/2020.acl-main.386/",
    doi = "10.18653/v1/2020.acl-main.386",
    pages = "4198--4205"
}

@article{lanham2023measuring,
  title={Measuring faithfulness in chain-of-thought reasoning},
  author={Lanham, Tamera and Chen, Anna and Radhakrishnan, Ansh and Steiner, Benoit and Denison, Carson and Hernandez, Danny and Li, Dustin and Durmus, Esin and Hubinger, Evan and Kernion, Jackson and others},
  journal={arXiv preprint arXiv:2307.13702},
  year={2023}
}

@inproceedings{mccoy2019right,
title = "Right for the Wrong Reasons: Diagnosing Syntactic Heuristics in Natural Language Inference",
    author = "McCoy, R. Thomas  and
      Pavlick, Ellie  and
      Linzen, Tal",
    editor = "Korhonen, Anna  and
      Traum, David  and
      M{\`a}rquez, Llu{\'i}s",
    booktitle = "Proceedings of the 57th Annual Meeting of the Association for Computational Linguistics",
    month = jul,
    year = "2019",
    address = "Florence, Italy",
    publisher = "Association for Computational Linguistics",
    url = "https://aclanthology.org/P19-1334/",
    doi = "10.18653/v1/P19-1334",
    pages = "3428--3448"
}

@inproceedings{chalkidis2022lexglue,
  title={{LexGLUE}: A Benchmark Dataset for Legal Language Understanding in {E}nglish},
  author={Chalkidis, Ilias and Jana, Abhik and Hartung, Dirk and Bommarito, Michael and Androutsopoulos, Ion and Katz, Daniel and Aletras, Nikolaos},
  booktitle={Proceedings of the 60th Annual Meeting of the Association for Computational Linguistics (Volume 1: Long Papers)},
  pages={4310--4330},
  year={2022}
}

@misc{rachmil2025whitening,
  title         = {Training-Free Policy Violation Detection via Activation-Space Whitening in {LLMs}},
  author        = {Rachmil, Oren and Shapira, Avishag and Betser, Roy and Hofman, Omer and Gershon, Itay and Shabtai, Asaf and Elovici, Yuval and Vainshtein, Roman},
  year          = {2025},
  eprint        = {2512.03994},
  archivePrefix = {arXiv},
  url           = {https://arxiv.org/abs/2512.03994}
}

@article{finguard2026,
  title={{FinGuard}: Detecting Financial Regulatory Non-Compliance in {LLM} Interactions},
  author={Dou, Huaixia and Zhu, Jie and Wu, Minghao and Jiang, Shuo and Li, Junhui and Guo, Lifan and Chen, Feng and Zhang, Chi},
  journal={arXiv preprint arXiv:2605.29427},
  year={2026}
}

@article{compliancenlp2026,
  title={{ComplianceNLP}: Knowledge-Graph-Augmented {RAG} for Multi-Framework Regulatory Gap Detection},
  author={Guo, Dongxin and Wu, Jikun and Yiu, Siu Ming},
  journal={arXiv preprint arXiv:2604.23585},
  year={2026}
}

@article{graphcompliance2025,
  title={{GraphCompliance}: Aligning Policy and Context Graphs for {LLM}-Based Regulatory Compliance},
  author={Chung, Jiseong and Ko, Ronny and Yoo, Wonchul and Onizuka, Makoto and Kim, Sungmok and Kim, Tae-Wan and Shin, Won-Yong},
  journal={arXiv preprint arXiv:2510.26309},
  year={2025}
}

@article{legilm2024,
  title={{LegiLM}: A Fine-Tuned Legal Language Model for Data Compliance},
  author={Zhu, Linkai and Yang, Lu and Li, Chaofan and Hu, Shanwen and Liu, Lu and Yin, Bin},
  journal={arXiv preprint arXiv:2409.13721},
  year={2024}
}

@inproceedings{judgingbytherules2026,
  title={Judging by the Rules: Compliance-Aligned Framework for Modern Slavery Statement Monitoring},
  author={Xu, Wenhao and Arodi, Akshatha and Nie, Jian-Yun and Fansi Tchango, Ars{\`e}ne},
  booktitle={Proceedings of the AAAI Conference on Artificial Intelligence},
  volume={40},
  pages={39477--39485},
  year={2026},
  doi={10.1609/aaai.v40i46.41298}
}

@inproceedings{notreadyforthebench2025,
title = "Not ready for the bench: {LLM} legal interpretation is unstable and uncalibrated to human judgments",
    author = "Purushothama, Abhishek  and
      Min, Junghyun  and
      Waldon, Brandon  and
      Schneider, Nathan",
    editor = "Aletras, Nikolaos  and
      Chalkidis, Ilias  and
      Barrett, Leslie  and
      Goanț{\u{a}}, C{\u{a}}t{\u{a}}lina  and
      Preoțiuc-Pietro, Daniel  and
      Spanakis, Gerasimos",
    booktitle = "Proceedings of the Natural Legal Language Processing Workshop 2025",
    month = nov,
    year = "2025",
    address = "Suzhou, China",
    publisher = "Association for Computational Linguistics",
    url = "https://aclanthology.org/2025.nllp-1.22/",
    doi = "10.18653/v1/2025.nllp-1.22",
    pages = "317",
    ISBN = "979-8-89176-338-8"
}

@misc{regops2026,
  title         = {Citation-Closure Retrieval and Per-Rule Attribution for Real-World Regulatory Compliance Question Answering},
  author        = {Ju, Yeong-Joon and Lee, Seong-Whan},
  year          = {2026},
  eprint        = {2605.29742},
  archivePrefix = {arXiv},
  url           = {https://arxiv.org/abs/2605.29742}
}

@inproceedings{clause2026,
  title={Better Call {CLAUSE}: A Discrepancy Benchmark for Auditing {LLMs} Legal Reasoning Capabilities},
  author={Choudhury, Manan Roy and Chandramouli, Adithya and Anand, Mannan and Gupta, Vivek},
  booktitle={Findings of the Association for Computational Linguistics: EACL 2026},
  pages={5776--5818},
  year={2026}
}

@inproceedings{gavel2026,
  title={{GAVEL}: Towards Rule-Based Safety through Activation Monitoring},
  author={Rozenfeld, Shir and Pankajakshan, Rahul and Zloczower, Itay and Lenga, Eyal and Gressel, Gilad and Mirsky, Yisroel},
  booktitle={The Fourteenth International Conference on Learning Representations},
  year={2026},
  url={https://openreview.net/forum?id=duntROHZ5R}
}

@article{refusalresidue2026,
  title={The Refusal Residue: {W}hen Probes Catch Alignment Faking and When They Don't},
  author={Mehta, Aman},
  journal={arXiv preprint arXiv:2607.13346},
  year={2026}
}

@inproceedings{highstakes2025,
  title={Detecting High-Stakes Interactions with Activation Probes},
  author={McKenzie, Alex and Pawar, Urja and Blandfort, Phil and Bankes, William and Krueger, David and Lubana, Ekdeep Singh and Krasheninnikov, Dmitrii},
  booktitle={Advances in Neural Information Processing Systems 38},
  pages={127556--127594},
  year={2025},
  doi={10.52202/085713-4249}
}

@inproceedings{legalbench2023guha,
author = {Guha, Neel and Nyarko, Julian and Ho, Daniel and R\'{e}, Christopher and Chilton, Adam and K, Aditya and Chohlas-Wood, Alex and Peters, Austin and Waldon, Brandon and Rockmore, Daniel and Zambrano, Diego and Talisman, Dmitry and Hoque, Enam and Surani, Faiz and Fagan, Frank and Sarfaty, Galit and Dickinson, Gregory and Porat, Haggai and Hegland, Jason and Wu, Jessica and Nudell, Joe and Niklaus, Joel and Nay, John and Choi, Jonathan and Tobia, Kevin and Hagan, Margaret and Ma, Megan and Livermore, Michael and Rasumov-Rahe, Nikon and Holzenberger, Nils and Kolt, Noam and Henderson, Peter and Rehaag, Sean and Goel, Sharad and Gao, Shang and Williams, Spencer and Gandhi, Sunny and Zur, Tom and Iyer, Varun and Li, Zehua},
booktitle = {Advances in Neural Information Processing Systems},
doi = {10.52202/075280-1915},
editor = {A. Oh and T. Naumann and A. Globerson and K. Saenko and M. Hardt and S. Levine},
pages = {44123--44279},
publisher = {Curran Associates, Inc.},
title = {{LegalBench}: A Collaboratively Built Benchmark for Measuring Legal Reasoning in Large Language Models},
url = {https://proceedings.neurips.cc/paper_files/paper/2023/file/89e44582fd28ddfea1ea4dcb0ebbf4b0-Paper-Datasets_and_Benchmarks.pdf},
volume = {36},
year = {2023}
}

@inproceedings{koreeda2021contractnli,
  title={{ContractNLI}: A Dataset for Document-level Natural Language Inference for Contracts},
  author={Koreeda, Yuta and Manning, Christopher D},
  booktitle={Findings of the Association for Computational Linguistics: EMNLP 2021},
  pages={1907--1919},
  year={2021}
}

@misc{sadhu2026compliancedetectorsreadaudit,
  title={What Do Compliance Detectors Read? {A}n Audit of Activation Probes and Guard Models},
  author={Saisab Sadhu and Aadit Sengupta and Vinay Kumar Sankarapu and Pratinav Seth},
  year={2026},
  eprint={2608.16852},
  archivePrefix={arXiv},
  primaryClass={cs.AI},
  url={https://arxiv.org/abs/2608.16852},
}

@inproceedings{sadhu2026rag,
  title={When {RAG} Disagrees: Detecting Latent Epistemic Conflict via Logit Interactions},
  author={Sadhu, Saisab and Roy, Dwaipayan and Basu, Tanmay},
  booktitle={Proceedings of the 49th International ACM SIGIR Conference on Research and Development in Information Retrieval},
  pages={4099--4104},
  year={2026},
  doi={10.1145/3805712.3809906}
}

@article{lpg2026,
  title={{LPG}: Balancing Efficiency and Policy Reasoning in Latent Policy Guardrails},
  author={Li, Nanxi and Zhao, Zhengyue and Xiao, Chaowei},
  journal={arXiv preprint arXiv:2605.17329},
  year={2026}
}

@article{rulesafevl2026,
  title={{RuleSafe-VL}: Evaluating Rule-Conditioned Decision Reasoning in Vision-Language Content Moderation},
  author={Lu, Zhifeng and Wang, Dianyuan and Shang, Yuhu and Xu, Zhenbo},
  journal={arXiv preprint arXiv:2605.07760},
  year={2026}
}

@inproceedings{dynaguard2026,
  title={{DynaGuard}: A Dynamic Guardian Model With User-Defined Policies},
  author={Hoover, Monte and Baherwani, Vatsal and Jain, Neel and Saifullah, Khalid and Vincent, Joseph and Jain, Chirag and Rad, Melissa Kazemi and Bruss, C. Bayan and Panda, Ashwinee and Goldstein, Tom},
  booktitle={The Fourteenth International Conference on Learning Representations},
  year={2026},
  url={https://openreview.net/forum?id=gc8Ylt0lbm}
}

@article{courtguard2026,
  title={{CourtGuard}: A Model-Agnostic Framework for Zero-Shot Policy Adaptation in {LLM} Safety},
  author={Suleymanov, Umid and Bayramov, Rufiz and Gafarli, Suad and Musayeva, Seljan and Mammadov, Taghi and Akhundlu, Aynur and Kantarcioglu, Murat},
  journal={arXiv preprint arXiv:2602.22557},
  year={2026}
}
\newpage
\appendix

\FloatBarrier
\section{Llama-Guard Prompt Adaptation}
\label{app:llamaguard}
Llama-Guard-3-8B's packaged chat template hardcodes 14 fixed
content-safety categories (S1: Violent Crimes ... S14: Code Interpreter
Abuse) with no field for an arbitrary policy. We bypass
\texttt{apply\_chat\_template} and construct the same delimited raw-text
scaffold the model was fine-tuned on directly, substituting a single
custom category built from the regulatory rule text for the fixed
S1-S14 block:

{\small\begin{verbatim}
<|begin_of_text|><|start_header_id|>user
<|end_header_id|>

Task: Check if there is a regulatory-
compliance violation in the 'User'
message below, according to the
compliance policy defined below.

<BEGIN COMPLIANCE POLICY>
O1: {rule text}
<END COMPLIANCE POLICY>

<BEGIN CONVERSATION>

User: {scenario text}

<END CONVERSATION>

Provide your compliance assessment
for ONLY THE LAST User message in
the above conversation:
 - First line must read 'safe' or
   'unsafe'.
 - If unsafe, a second line must
   include a comma-separated list
   of violated categories.
<|eot_id|><|start_header_id|>assistant
<|end_header_id|>
\end{verbatim}
}
\texttt{safe} is read as compliant (1), \texttt{unsafe} as non-compliant (0).

\FloatBarrier
\section{Domains and OCS Audit Statistics}
\label{app:domains}
Table~\ref{tab:domains} gives the exact 20 domains used across all
experiments, total examples per domain in OmniCompliance-100K, and the
fraction of that
domain's own per-example rule texts containing at least one mandatory
modal (``must''/``shall''), which determines OCS-neg applicability
(Section~\ref{sec:conditions}). This gate applies per sample, not per
domain, so a domain's OCS-neg sample count is its sampled N times this
rate rather than an all-or-nothing cutoff, except for the three domains
at exactly 0.0\%.

\begin{table}[t]
\centering
\caption{Domain sizes and per-sample mandatory-modal coverage (OCS-neg
applicability). The three domains at 0.0\% (MITRE ATT\&CK technique
descriptions, academic-integrity value statements, and online-learning
guidance) are not phrased in mandatory-obligation language at all. This
is a genuine linguistic property of these domains, not a processing
failure%
; with nothing to negate, they are excluded from OCS-neg entirely. $^\dagger$Domain name as in OmniCompliance-100K.}
\vspace{3mm}
\label{tab:domains}
\footnotesize
\setlength{\tabcolsep}{6pt}
\renewcommand{\arraystretch}{1.1}
\begin{tabular}{@{}lrr@{}}
\toprule
Domain & $n$ total & Modal cov. \\
\midrule
gdpr & 3{,}759 & 59.1\% \\
hipaa & 6{,}924 & 20.5\% \\
eu\_ai\_act & 7{,}758 & 60.7\% \\
ccpa & 2{,}489 & 30.9\% \\
sb35 & 1{,}244 & 31.8\% \\
finance\_crypto & 1{,}969 & 38.9\% \\
finance\_anti\_laundering\_and\_terrorist & 4{,}875 & 50.4\% \\
finance\_cross\_border\_payment\_law & 619 & 46.4\% \\
finance\_eletric\_momey$^\dagger$ & 1{,}117 & 43.6\% \\
data\_act & 4{,}415 & 49.5\% \\
cybersecurity\_mitre\_attack & 513 & 0.0\% \\
edu\_academic\_integrity & 414 & 0.0\% \\
edu\_online\_learning & 757 & 0.0\% \\
edu\_discrimination\_us\_edu\_dept & 146 & 22.6\% \\
policy\_github & 9{,}159 & 10.4\% \\
policy\_google & 4{,}855 & 4.3\% \\
policy\_openai & 1{,}091 & 10.1\% \\
policy\_reddit & 6{,}785 & 8.0\% \\
policy\_x & 1{,}143 & 9.5\% \\
foundation\_rights & 1{,}822 & 34.5\% \\
\bottomrule
\end{tabular}
\end{table}

Every domain reaches the target $N=200$ per model except
\texttt{edu\_discrimination\_us\_edu\_dept}, which has only 146 examples
total in OmniCompliance-100K and uses all of them, with no oversampling.
This is consistent across all five models, since sampling is seeded by
(domain, $N$, seed) alone.

\begin{figure}[h]
\centering
\includegraphics[width=0.4664\columnwidth]{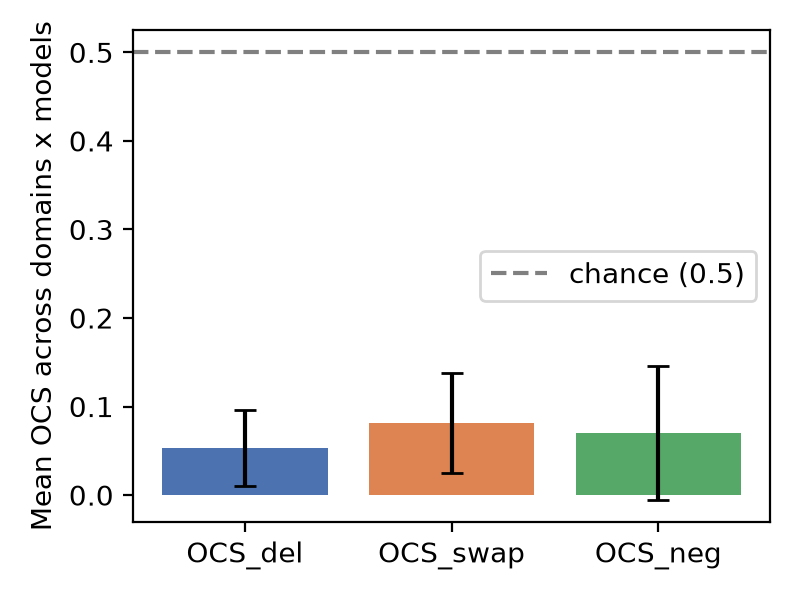}
\caption{Mean OCS by perturbation condition, pooled across all models
and domains. OCS-del (rule deleted) is the lowest of the three:
verdicts are least likely to change when the rule is removed entirely.}
\label{fig:perturbation-bars}
\end{figure}

Table~\ref{tab:accuracy-perturbation} gives baseline accuracy and
accuracy under each rule perturbation condition, and
Figure~\ref{fig:perturbation-bars} gives the per-condition breakdown
underlying the pooled OCS-del/OCS-swap/OCS-neg numbers, both summarized
in Section~\ref{sec:results-audit}.

\begin{table}[h]
\centering
\caption{Baseline accuracy and accuracy under each rule perturbation
condition, pooled across all 20 domains. Accuracy under rule deletion
approximates performance when only case content is available.}
\vspace{3mm}
\label{tab:accuracy-perturbation}
\footnotesize
\setlength{\tabcolsep}{8pt}
\renewcommand{\arraystretch}{1.2}
\begin{tabular}{@{}lcccc@{}}
\toprule
Model & Baseline & Del & Swap & Neg \\
\midrule
Qwen2.5-7B & 90.3\% & 90.7\% & 83.2\% & 79.5\% \\
Llama-3.1-8B & 91.9\% & 91.4\% & 89.4\% & 89.7\% \\
Llama-3.3-70B & 90.9\% & 87.5\% & 84.5\% & 83.6\% \\
Llama-Guard-3-8B & 51.1\% & 50.0\% & 50.1\% & 50.9\% \\
Mistral-7B-v0.3 & 90.3\% & 91.6\% & 89.6\% & 88.7\% \\
\bottomrule
\end{tabular}
\end{table}

\FloatBarrier
\section{Prompt Templates}
\label{app:prompts}
Standard verdict-elicitation prompt (all Tier-1 general models):

{\small\begin{verbatim}
Rule: {rule text, first 300 chars}

Scenario: {case background}

Question: Does this scenario comply
with the stated rule? Answer with
exactly one word: Compliant or
Non-compliant.

Answer:
\end{verbatim}
}

Rule-grounded prompt (Section~\ref{sec:interventions}):

{\small\begin{verbatim}
Rule: {rule text}

Scenario: {case background}

Instruction: First identify and copy
the specific clause of the rule that
applies most directly to this
scenario. Then, on a new line
starting with exactly 'Compliance
verdict:', answer with exactly one
word: Compliant or Non-compliant.

Most relevant clause:
\end{verbatim}
}

Verdicts are parsed by checking for ``non-compliant''/``non compliant''
before ``compliant'', so a compliant-then-negated phrasing is not
misread. If neither string appears, the verdict is returned as null and
excluded from that condition's OCS rather than coerced into a guess.

\FloatBarrier
\section{ICS-delta and Convergent Validity}
\label{app:convergent-validity}
\begin{figure}[h]
\centering
\includegraphics[width=0.4664\columnwidth]{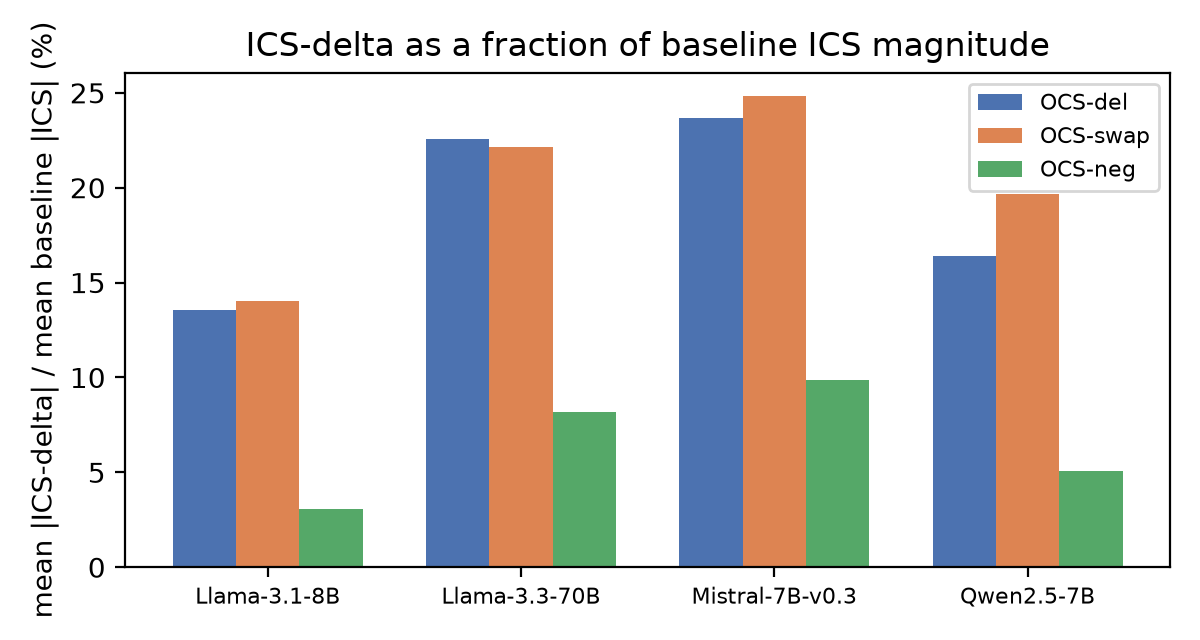}
\caption{ICS-delta as a percentage of baseline ICS magnitude, per model
per condition. Every bar is a small fraction of the baseline signal;
OCS-neg (rightmost, green) is consistently the smallest.}
\label{fig:ics-delta-bars}
\end{figure}

Table~\ref{tab:ics-delta} and Figure~\ref{fig:ics-delta-bars} give the
mean baseline ICS magnitude, mean $|$ICS-delta$|$, and their ratio per
Tier-1 model, summarized in Section~\ref{sec:ics-delta}.

\begin{table}[h]
\centering
\caption{Mean baseline ICS magnitude, mean $|$ICS-delta$|$, and their
ratio, pooled across all three perturbation conditions and 20 domains.}
\vspace{3mm}
\label{tab:ics-delta}
\footnotesize
\setlength{\tabcolsep}{8pt}
\renewcommand{\arraystretch}{1.2}
\begin{tabular}{@{}lccc@{}}
\toprule
Model & mean $|$ICS$|$ & mean $|$ICS-delta$|$ & Ratio \\
\midrule
Llama-3.1-8B & 4.56 & 0.58 & 12.8\% \\
Llama-3.3-70B & 3.31 & 0.69 & 21.0\% \\
Mistral-7B-v0.3 & 1.45 & 0.33 & 22.9\% \\
Qwen2.5-7B & 16.71 & 2.79 & 16.7\% \\
\bottomrule
\end{tabular}
\end{table}

Table~\ref{tab:convergent-validity-appendix} gives the per-model Spearman
correlation between OCS-agg and ICS-delta-agg rankings across all 20
domains, summarized in Section~\ref{sec:convergent-validity}.

\begin{table}[h]
\centering
\caption{OCS-agg vs.\ ICS-delta-agg per-domain rank correlation, one row
per Tier-1 model ($n=20$ domains each). None reaches conventional
significance.}
\vspace{3mm}
\label{tab:convergent-validity-appendix}
\footnotesize
\setlength{\tabcolsep}{8pt}
\renewcommand{\arraystretch}{1.2}
\begin{tabular}{@{}lcc@{}}
\toprule
Model & $\rho$ & $p$ \\
\midrule
Qwen2.5-7B-Instruct & 0.114 & 0.631 \\
Llama-3.1-8B-Instruct & 0.084 & 0.724 \\
Mistral-7B-Instruct-v0.3 & $-0.032$ & 0.895 \\
Llama-3.3-70B-Instruct & 0.338 & 0.145 \\
\bottomrule
\end{tabular}
\end{table}

\FloatBarrier
\section{Empirical Null and Rule-Necessity}
\label{app:necessity}
Table~\ref{tab:permutation-null} gives the full per-model bias-corrected
null, observed OCS-agg, and permutation $p$-value summarized in
Section~\ref{sec:empirical-null}.

\begin{table}[h]
\centering
\caption{Observed OCS-agg vs.\ an independence null built from each
model's own empirical baseline/altered verdict marginals
($p(1-q){+}q(1-p)$), validated by a 10,000-sample within-domain
permutation test ($p$: one-sided, observed $\leq$ null). Every model's
OCS is significantly below its own bias-corrected null, not only below
the naive 0.5 reference.}
\vspace{3mm}
\label{tab:permutation-null}
\footnotesize
\setlength{\tabcolsep}{8pt}
\renewcommand{\arraystretch}{1.2}
\begin{tabular}{@{}lccc@{}}
\toprule
Model & Null (indep.) & Obs.\ OCS & Perm.\ $p$ \\
\midrule
Llama-3.1-8B & 0.500 & 0.070 & $<0.0001$ \\
Llama-3.3-70B & 0.486 & 0.081 & $<0.0001$ \\
Mistral-7B-v0.3 & 0.502 & 0.089 & $<0.0001$ \\
Qwen2.5-7B & 0.483 & 0.094 & $<0.0001$ \\
Llama-Guard-3-8B & 0.019 & 0.013 & $<0.0001$ \\
\bottomrule
\end{tabular}
\end{table}
Table~\ref{tab:necessity-appendix} gives OCS-swap and OCS-neg on each
model's own rule-necessary subset (Section~\ref{sec:necessity-control}),
with both one-sided permutation $p$-values against that model's own
bias-corrected null: $p_\downarrow$ tests whether the observed rate is
significantly below the null (the direction relevant to the paper's
main claim), $p_\uparrow$ tests significantly above (the direction that
turns out to hold for OCS-swap). Both use the same permutation-null
procedure as Table~\ref{tab:permutation-null}, but with 20,000
permutations per cell here rather than 10,000, for finer resolution at
the extreme tail values this table reports.

\begin{table}[t]
\centering
\caption{Full rule-necessity control statistics: OCS-swap and OCS-neg
per model, restricted to that model's own rule-necessary subset
(baseline correct, del incorrect). Llama-3.3-70B-Instruct's nominal
$p_\downarrow=0.036$ for OCS-neg does not survive any correction for
the ten tests conducted here (Bonferroni threshold $0.005$); we do not
treat it as a significant reduction.}
\vspace{3mm}
\label{tab:necessity-appendix}
\footnotesize
\setlength{\tabcolsep}{6pt}
\renewcommand{\arraystretch}{1.2}
\begin{tabular}{@{}llccccc@{}}
\toprule
Model & Cond. & $n$ & Obs. & Null & $p_\downarrow$ & $p_\uparrow$ \\
\midrule
Llama-3.1-8B & swap & 132 & 0.780 & 0.669 & 1.000 & $<0.0001$ \\
Llama-3.1-8B & neg & 42 & 0.333 & 0.437 & 0.081 & 0.987 \\
Llama-3.3-70B & swap & 187 & 0.733 & 0.723 & 0.881 & 0.329 \\
Llama-3.3-70B & neg & 53 & 0.377 & 0.479 & 0.036 & 0.996 \\
Mistral-7B-v0.3 & swap & 120 & 0.750 & 0.665 & 0.999 & 0.006 \\
Mistral-7B-v0.3 & neg & 36 & 0.389 & 0.486 & 0.166 & 0.961 \\
Qwen2.5-7B & swap & 94 & 0.745 & 0.695 & 0.994 & 0.050 \\
Qwen2.5-7B & neg & 31 & 0.516 & 0.552 & 0.492 & 0.797 \\
Llama-Guard-3-8B & swap & 49 & 1.000 & 1.000 & 1.000 & 1.000 \\
Llama-Guard-3-8B & neg & 23 & 0.174 & 0.174 & 1.000 & 1.000 \\
\bottomrule
\end{tabular}
\end{table}

\FloatBarrier
\section{OCS as a Predictor of Accuracy}
\label{app:accuracy-predictor}
\begin{figure}[h]
\centering
\includegraphics[width=0.4664\columnwidth]{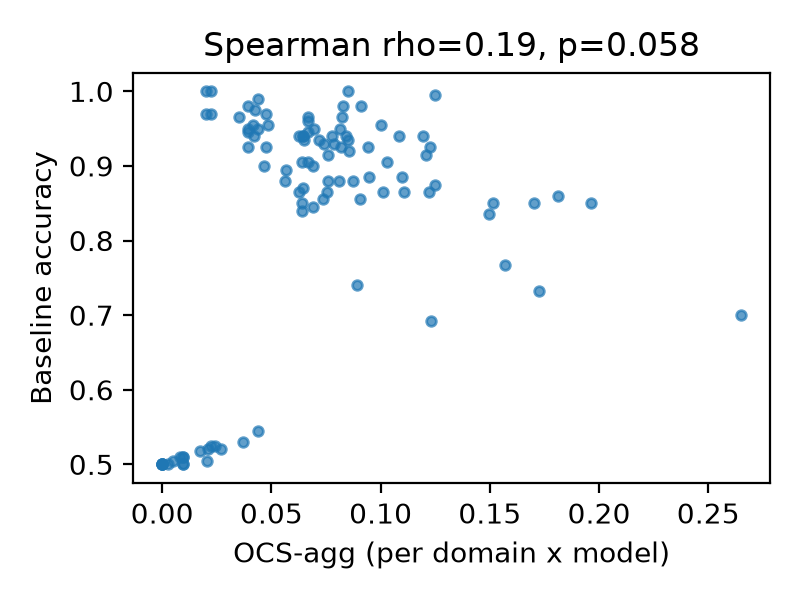}
\caption{OCS-agg vs.\ baseline accuracy, all 100 (model, domain) cells.
The correlation is positive but narrowly misses $p<0.05$.}
\label{fig:ocs-vs-acc}
\end{figure}

Spearman correlation between OCS-agg and baseline accuracy across all
100 (model, domain) cells is weak and narrowly misses conventional
significance ($\rho = 0.190$, $p = 0.058$, $n=100$;
Figure~\ref{fig:ocs-vs-acc}). We treat this as a marginal relationship:
OCS-agg alone would not reliably substitute for gold-label accuracy
measurement at this sample size, though the direction is consistent with
higher rule-sensitivity predicting higher accuracy.

\FloatBarrier
\section{Linguistic Regression: Full Statistics}
\label{app:linguistic}
\begin{figure}[h]
\centering
\includegraphics[width=0.4664\columnwidth]{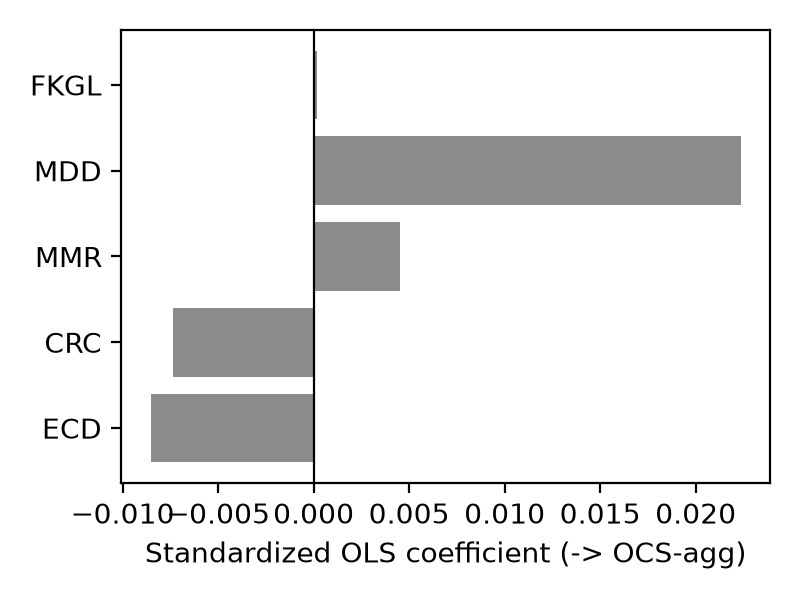}
\caption{Standardized OLS coefficients (all five features jointly,
$\to$ OCS-agg). Red bars mark $p<0.05$ individually within the joint
model; none reach this threshold, and MDD is closest.}
\label{fig:regression-coefs}
\end{figure}

Table~\ref{tab:regression} gives the per-feature correlations, and
Figure~\ref{fig:regression-coefs} the joint standardized OLS
coefficients, summarized in Section~\ref{sec:linguistic}.

\begin{table}[t]
\centering
\caption{Linguistic feature correlations with per-domain mean OCS-agg
($n=19$--20 domains). Only MDD survives Benjamini-Hochberg correction.}
\vspace{3mm}
\label{tab:regression}
\footnotesize
\setlength{\tabcolsep}{8pt}
\renewcommand{\arraystretch}{1.2}
\begin{tabular}{@{}lccc@{}}
\toprule
Feature & Spearman $\rho$ & $p_{BH}$ & Sig.\ (BH 0.05) \\
\midrule
ECD & 0.041 & 0.86 & no \\
CRC & 0.241 & 0.38 & no \\
MMR & 0.405 & 0.14 & no \\
MDD & 0.570 & 0.044 & \textbf{yes} \\
FKGL & 0.448 & 0.12 & no \\
\bottomrule
\end{tabular}
\end{table}

\FloatBarrier
\section{Interventions: Full Statistics}
\label{app:steering}
Table~\ref{tab:exp3a} gives the rule-grounded prompting statistics
summarized in Section~\ref{sec:rule-grounded}.

\begin{table}[H]
\centering
\caption{Rule-grounded prompting: accuracy change vs.\ standard
prompting on the LOW/HIGH-OCS partitions (Llama-3.3-70B-Instruct's
per-sample OCS), parseable-verdict cases only; $p$ is McNemar's exact
test on LOW. Only Qwen2.5-7B reaches significance, and harmfully.
$^\dagger$Robustness check, not a grounded-reasoning test
(Section~\ref{sec:models}).}
\vspace{3mm}
\label{tab:exp3a}
\footnotesize
\setlength{\tabcolsep}{8pt}
\renewcommand{\arraystretch}{1.2}
\begin{tabular}{@{}lccc@{}}
\toprule
Model & $\Delta_{\text{low}}$ & $\Delta_{\text{high}}$ & $p$ (LOW) \\
\midrule
Llama-3.1-8B & $-0.006$ & $0.000$ & $0.146$ \\
Llama-3.3-70B & $+0.007$ & $-0.033$ & $0.073$ \\
Qwen2.5-7B & $\mathbf{-0.028}$ & $-0.006$ & $\mathbf{<0.0001}$ \\
Mistral-7B-v0.3 & $-0.010$ & $0.000$ & $0.077$ \\
Llama-Guard-3-8B$^\dagger$ & $-0.001$ & $0.000$ & $0.500$ \\
\bottomrule
\end{tabular}
\end{table}

\begin{figure}[H]
\centering
\includegraphics[width=0.65\columnwidth]{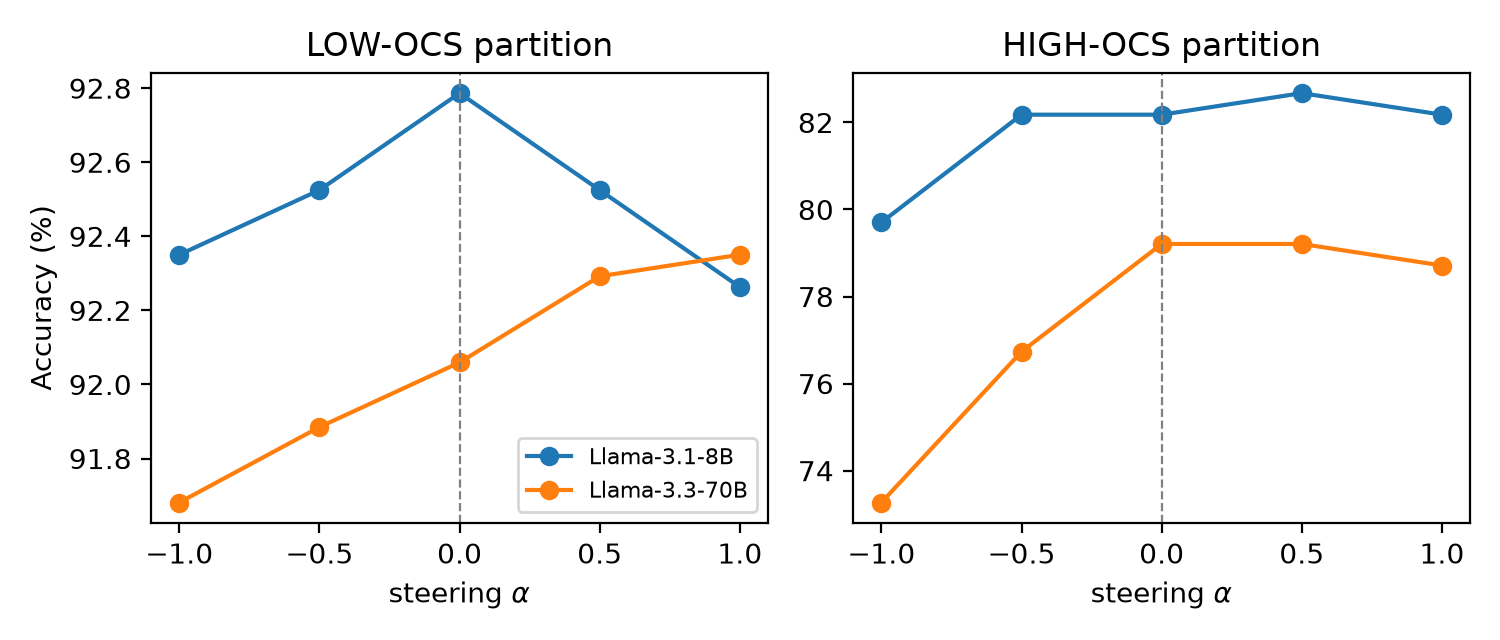}
\caption{Accuracy vs.\ steering $\alpha$, LOW- and HIGH-OCS partitions,
both models. Negative $\alpha$ degrades accuracy for both models on both
partitions; positive $\alpha$ never significantly helps.}
\label{fig:steering-sweep}
\end{figure}

Table~\ref{tab:steering-appendix} gives the exact McNemar's-test $p$-value
for every (model, partition, $\alpha$) cell summarized in
Table~\ref{tab:steering}, and Figure~\ref{fig:steering-sweep} plots the
same sweep, each tested against that model and partition's unsteered
($\alpha=0$) baseline.

\begin{table}[H]
\centering
\caption{Exact McNemar's-test $p$-values, each (model, partition,
$\alpha$) cell vs.\ that row's $\alpha=0$ baseline, paired per case.}
\vspace{3mm}
\label{tab:steering-appendix}
\footnotesize
\setlength{\tabcolsep}{8pt}
\renewcommand{\arraystretch}{1.2}
\begin{tabular}{@{}llcccc@{}}
\toprule
Model & Part. & $\alpha=-1$ & $-0.5$ & $+0.5$ & $+1$ \\
\midrule
Llama-3.1-8B & LOW & 0.0489 & 0.1628 & 0.1496 & 0.0273 \\
Llama-3.1-8B & HIGH & 0.2668 & 1.0000 & 1.0000 & 1.0000 \\
Llama-3.3-70B & LOW & 0.0044 & 0.1094 & 0.1153 & 0.1742 \\
Llama-3.3-70B & HIGH & 0.0290 & 0.3323 & 1.0000 & 1.0000 \\
\bottomrule
\end{tabular}
\end{table}

\end{document}